\documentclass[letterpaper,twoside]{article}
\usepackage[preprint]{aistats2027}
\usepackage[utf8]{inputenc}
\usepackage[T1]{fontenc}
\usepackage[round]{natbib}
\usepackage{amsmath,amssymb}
\usepackage{booktabs,array,graphicx,microtype,needspace}
\usepackage{xcolor,tikz}
\usetikzlibrary{arrows.meta,positioning,calc,fit}
\usepackage{algorithm,algpseudocode}
\usepackage{placeins,balance}
\usepackage[hidelinks]{hyperref}
\hypersetup{pdftitle={Learning Graph Energies for Finite-Time Generation},pdfauthor={Amartya Banerjee, Xingyu Xu, Caroline Moosmüller, Harlin Lee}}
\newcommand{\VG}{V_{\mathrm G}}
\newcommand{\Vpre}{V_{\mathrm{pre}}}
\newcommand{\bref}{\beta_{\mathrm{ref}}}
\newcommand{\E}{\mathbb E}

\newcommand{\TV}{\operatorname{TV}}
\newcommand{\KL}{\operatorname{KL}}

\newcommand{\papertitle}{Learning Graph Energies for Finite-Time Generation}
\begin{document}
\runningauthor{Amartya Banerjee, Xingyu Xu, Caroline Moosmüller, Harlin Lee}
\brokenpenalty=10000
\setlength{\emergencystretch}{1em}
\twocolumn[
\aistatstitle{\papertitle}
\aistatsauthor{%
Amartya Banerjee \And
Xingyu Xu \And
Caroline Moosmüller \And
Harlin Lee}
\aistatsaddress{%
\footnotesize Dept.\ of Computer Science\\ \footnotesize UNC Chapel Hill \And
\footnotesize Carnegie Mellon University \And
\footnotesize Dept.\ of Mathematics\\ \footnotesize UNC Chapel Hill \And
\footnotesize SDSS\\ \footnotesize UNC Chapel Hill}
]
\begin{abstract}
Energy-based models can represent distributions over discrete structures such as molecular graphs, but an energy that is accurate at equilibrium need not yield accurate samples when the sampler stops after a fixed number of steps. We introduce finite-horizon energy learning, which adapts the acceptance energy of a Metropolis--Hastings sampler to the distribution it returns after a prescribed number of steps, with initialization, proposals, and annealing schedule held fixed. For graph generation, a frozen pretrained model supplies initialization and proposals, and a learned energy governs acceptance. A training-only discriminator scores completed samples, and coupled continuations of the accepted and rejected branches of a proposal give an acceptance-gradient estimator that is unbiased for a fixed discriminator under stated conditions, without equilibrated negative samples or differentiation through discrete states. Generation uses the learned energy in the original single-trajectory sampler, without discriminator evaluations or reranking. On twelve pairwise binary targets with exactly evaluated terminal laws, learned energies have lower median terminal total variation than the equilibrium-correct energy in every tested short-horizon setting, and than its numerically tuned scalar multiple in most settings. On MOSES, a learned two-coefficient combination of pretrained Graph Energy Matching (GEM) potentials lowers FCD by 3.7--10.5\% relative to scale-calibrated GEM at matched sampling steps, with validity gains at the shortest budget and a sub-point validity trade-off at higher budgets.
\end{abstract}
\vfill
\section{Introduction}
\label{sec:intro}

\begin{figure*}[t]
\centering
\resizebox{\textwidth}{!}{\definecolor{ovBlue}{HTML}{0072B2}
\begin{tikzpicture}[
 x=1cm,y=1cm,font=\fontsize{8}{9.5}\selectfont,
 >={Stealth[length=1.8mm,width=1.2mm]},
 every node/.style={inner sep=3pt},
 frozen/.style={draw=black!60,fill=black!6,rounded corners=2pt,line width=.6pt,align=center},
 learned/.style={draw=ovBlue!85!black,fill=ovBlue!8,rounded corners=2pt,line width=.85pt,align=center},
 training/.style={draw=black!60,fill=white,dashed,rounded corners=2pt,line width=.65pt,align=center},
 flow/.style={->,line width=.8pt,shorten <=1pt,shorten >=1pt},
 trainline/.style={dashed,draw=black!65,line width=.7pt,rounded corners=3pt},
 trainflow/.style={trainline,->,shorten >=1pt},
 updateflow/.style={->,draw=ovBlue!85!black,dashed,line width=.85pt,rounded corners=4pt,shorten >=1pt},
 tag/.style={font=\fontsize{7}{8.5}\selectfont\bfseries,inner sep=1pt}]

\draw[|-|,draw=black!65,line width=.5pt] (0,5.71)--(14.05,5.71);
\node[anchor=south,inner sep=2pt] at (7.025,5.71)
 {\textbf{Generation}: one sampling trajectory};
\draw[|-|,dashed,draw=black!65,line width=.5pt] (14.35,5.71)--(16.85,5.71);
\node[anchor=south,inner sep=2pt] at (15.6,5.71) {\textbf{Training only}};

\draw[draw=black!55,rounded corners=3pt,line width=.55pt]
 (3.15,3.20) rectangle (12.03,5.48);
\node[inner sep=0pt] at (7.59,5.27) {$K$ Metropolis--Hastings steps};

\node[frozen,text width=1.90cm,minimum height=1.28cm] (warm) at (1.04,4.10)
 {Initialization\\from noise\\using $\VG$};
\node[tag,anchor=south west] at (warm.north west) {frozen};
\node[draw=black!65,circle,inner sep=1.5pt] (x0) at (2.70,4.10) {$X_0$};
\node[frozen,text width=2.70cm,minimum height=1.28cm] (proposal) at (4.85,4.10)
 {Propose an edit\\$Y_t\sim Q_t(\cdot\mid X_t)$\\using $\VG$};
\node[tag,anchor=south west] at (proposal.north west) {frozen};
\node[learned,text width=4.40cm,minimum height=1.28cm] (accept) at (9.45,4.10)
 {Accept $Y_t$ with probability $\alpha_{\theta,t}$\\computed from the learned\\acceptance energy $E_\theta$};
\node[tag,anchor=south west,text=ovBlue!70!black] at (accept.north west) {learned: $\theta$};
\node[draw=black!60,rounded corners=2pt,text width=1.55cm,align=center,minimum height=1.28cm]
 (output) at (13.10,4.10) {Completed\\graph $X_K$};
\node[training,text width=2.25cm,minimum height=1.28cm] (critic) at (15.60,4.10)
 {Discriminator $d_\psi$\\terminal reward\\$r_\psi(X_K)$};
\draw[flow] (warm.east)--(x0.west);
\draw[flow] (x0.east)--(proposal.west);
\draw[flow] (proposal.east)--(accept.west);
\draw[flow] (accept.east)--(output.west);
\draw[trainflow] (output.east)--(critic.west);

\node[anchor=west,inner sep=0pt] at (4.10,2.75)
 {\textbf{Training}: compare accept and reject continuations};
\node[training,circle,minimum size=.50cm,inner sep=0pt] (fork) at (3.30,1.32) {$t$};
\coordinate (selectiontop) at (3.30,3.00);
\draw[densely dotted,draw=black!60,line width=.7pt,rounded corners=3pt]
 (accept.south west)--(accept.south west |- selectiontop)--(selectiontop)--(fork.north);

\node[training,text width=2.75cm,minimum height=.78cm] (yes) at (5.95,1.94)
 {Accept: start at $Y_t$\\Continue to step $K$};
\node[training,text width=2.75cm,minimum height=.78cm] (no) at (5.95,.70)
 {Reject: start at $X_t$\\Continue to step $K$};
\node[training,text width=1.35cm,minimum height=.78cm] (rplus) at (8.62,1.94) {reward $R^+$};
\node[training,text width=1.35cm,minimum height=.78cm] (rminus) at (8.62,.70) {reward $R^-$};

\coordinate (branch) at (3.99,1.32);
\draw[trainline] (fork.east)--(branch);
\draw[trainflow] (branch)|-(yes.west);
\draw[trainflow] (branch)|-(no.west);
\draw[trainflow] (yes.east)--(rplus.west);
\draw[trainflow] (no.east)--(rminus.west);
\coordinate (rewardjoin) at (9.85,1.32);
\draw[trainline] (rplus.east)-|(rewardjoin);
\draw[trainline] (rminus.east)-|(rewardjoin);
\node[learned,text width=6.30cm,minimum height=1.56cm] (gradient) at (13.45,1.32)
 {Update the acceptance-energy coefficients\\[4pt]
 $\displaystyle\widehat g=\frac1N\sum_{i,t}\frac{I_{it}}{p_{it}}(R_{it}^+-R_{it}^-)\nabla_\theta\alpha_{\theta,t}$\\[4pt]
 Unbiased for $\nabla_\theta J$ at fixed $d_\psi$ (Section~\ref{sec:estimator})};
\draw[trainflow] (rewardjoin)--(gradient.west);

\coordinate (feedback) at (11.42,2.80);
\draw[updateflow] (gradient.north)--(gradient.north |- feedback)--(feedback)--(accept.south -| feedback);
\node[anchor=south,font=\fontsize{7.2}{8.5}\selectfont,inner sep=1pt,text=ovBlue!70!black]
 at (12.46,2.84) {update $\theta$};
\node[anchor=north,inner sep=1pt] at (5.95,.14) {Shared future random numbers};

\node[frozen,minimum width=.50cm,minimum height=.28cm] (key1) at (.43,1.98) {};
\node[anchor=west] at (key1.east) {frozen};
\node[learned,minimum width=.50cm,minimum height=.28cm] (key2) at (.43,1.32) {};
\node[anchor=west] at (key2.east) {learned};
\node[training,minimum width=.50cm,minimum height=.28cm] (key3) at (.43,.66) {};
\node[anchor=west] at (key3.east) {training only};
\end{tikzpicture}}
\caption{\textbf{Finite-horizon energy learning for a pretrained GEM graph sampler.} Generation (top): one $K$-step MH trajectory whose initialization and proposals use GEM's frozen potential $\VG$; only the acceptance energy $E_\theta$, with coefficients $\theta=(c,\omega)$ in \eqref{eq:energy}, is learned. Training (dashed; absent at generation): at a selected decision $t$, the accepted and rejected branches are continued to step $K$ with shared future random numbers, and the difference of their terminal rewards weights $\nabla_\theta\alpha_{\theta,t}$. The displayed update is \eqref{eq:estimator} with selection indicators $I_{it}$ and inclusion probabilities $p_{it}$; $N$ is the number of base chains.}
\label{fig:method}
\end{figure*}
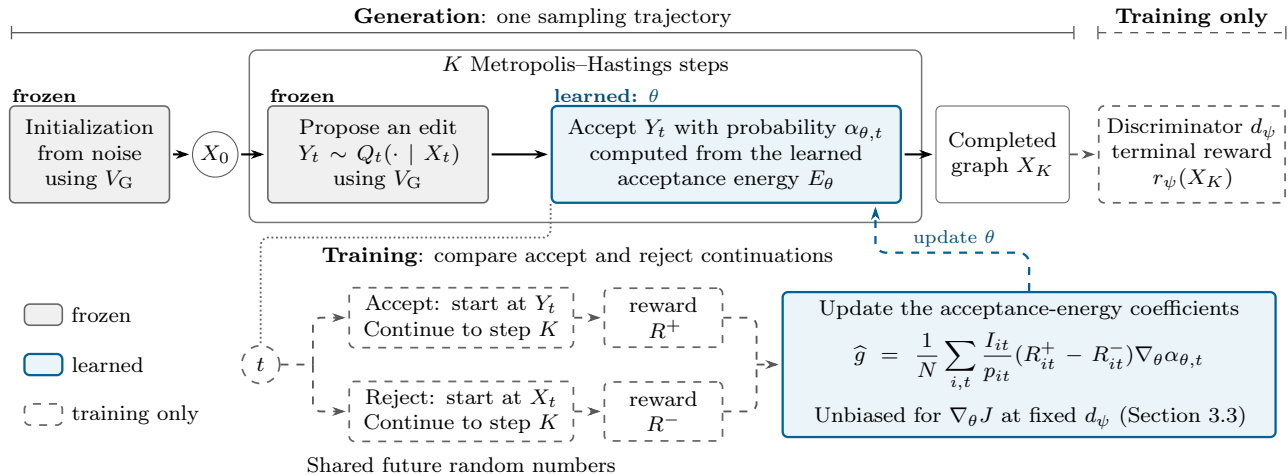

For a generator stopped after $K$ transitions, sampling accuracy is determined by its terminal distribution, not by its equilibrium law. The terminal distribution depends on initialization, proposals, acceptance decisions, and the annealing schedule. Consequently, even an energy that represents the target exactly at equilibrium need not minimize error after a finite run. Prior work on short-run energy-based models establishes the importance of this distinction \citep{nijkamp2019learning,agoritsas2023explaining}.

We study whether terminal sample feedback can improve finite-run accuracy by changing the acceptance energy alone, while retaining the initialization, proposal kernel, and schedule. A two-state chain illustrates why this can help (Appendix~\ref{app:example}). Let the target be uniform on $\{a,b\}$, start the chain at $a$, and propose a flip with probability $3/4$. At inverse temperature one, the correct energy accepts every flip and returns $b$ with probability $3/4$ after one step. Raising the energy of $b$ by $\log(3/2)$ lowers the acceptance probability to $2/3$, making the one-step output exactly uniform but moving the equilibrium probability of $b$ to $2/5$.

We introduce \emph{finite-horizon energy learning},\footnote{Code is available at \url{https://github.com/amartya21/Finite-Time-Graph-Sampling.git}.} which optimizes the acceptance energy using discriminator rewards on completed $K$-step samples. For pretrained graph generators such as Graph Energy Matching (GEM; \citealp{balcerak2026gem}), which refines graphs with Metropolis--Hastings (MH) steps using gradient-informed proposals from a learned potential \citep{zhang2022langevin}, this separates the frozen potential used for initialization and proposals from the energy adapted for acceptance. An acceptance decision changes the state from which subsequent proposals are drawn, so its effect must be assessed through the remaining trajectory.

Following the coupled MH derivative of \citet{arya2023differentiating,arya2024gradient}, we continue both the accepted and rejected branches of a selected decision to the terminal horizon using shared random numbers. Their terminal-reward difference weights the derivative of the local acceptance probability. The discriminator and additional continuations are used only during training; generation uses the learned acceptance energy in a single trajectory (Figure~\ref{fig:method}).

The finite-state study isolates acceptance-energy adaptation from density misspecification by comparing against a target energy that is already correct at equilibrium, and against a numerically tuned scalar multiple of that energy. The molecular study then applies the same terminal-reward objective to a two-coefficient combination of frozen GEM potentials on MOSES \citep{polykovskiy2020moses}. Our contributions are:
\begin{itemize}
\item \textbf{Finite-horizon optimization of acceptance energies.} We formulate learning through the MH acceptance energy to maximize expected terminal discriminator reward, with initialization, the proposal kernel, and the annealing schedule fixed. Applying established coupled MH derivatives to subsampled accept/reject continuations gives an unbiased gradient estimator for a fixed discriminator under the conditions in Section~\ref{sec:estimator}.
\item \textbf{Exact finite-state evidence beyond scalar tempering.} At $K\in\{4,8,16\}$, learned energies achieve lower median terminal total variation (TV) than the equilibrium-correct energy in all 18 reference and all 18 new-target settings. They also outperform its numerically tuned scalar multiple in 14 and 15 settings, respectively. These comparisons identify finite-horizon improvements beyond the reported scalar-temperature control in most settings (Section~\ref{sec:exact}).
\item \textbf{Adaptation of frozen graph potentials.} On MOSES, a two-coefficient acceptance energy combining pretrained GEM potentials reduces Fr\'echet ChemNet distance (FCD) by 3.7--10.5\% relative to scale-calibrated GEM with exact MH acceptance at matched refinement steps, with budget-dependent validity trade-offs. Generation retains a single trajectory and requires neither discriminator evaluations nor additional continuations (Section~\ref{sec:molecules}).
\end{itemize}

\section{Related Work}
\label{sec:related}

\paragraph{Training energies with finite-run samplers.}
Short-run EBMs treat a fixed number of MCMC steps as part of the generator \citep{nijkamp2019learning,agoritsas2023explaining}, and contrastive divergence and GEM's contrastive refinement raise energy on generated negatives held fixed during differentiation \citep{hinton2002training,balcerak2026gem}. These methods learn the energy, and with it the sampler's target, from short-run negatives. We instead fix a pretrained sampler's initialization and proposals and learn only its acceptance energy, for an objective defined on the $K$-step output.

\paragraph{Adapting proposals and transition dynamics.}
ALOE and EB-GFN learn discrete energies jointly with amortized samplers \citep{dai2020aloe,zhang2022ebgfn}, A-NICE-MC trains MCMC proposals adversarially \citep{song2017anicemc}, and reinforcement learning adapts MCMC kernels \citep{wang2025reinforcement}. For graphs, DeFoG separates the trained denoiser from the choice of sampling dynamics \citep{qin2025defog}. These approaches change how states are proposed or how the sampler moves; we keep the proposal kernel and schedule fixed and change only the acceptance energy.

\paragraph{Differentiating through discrete randomness.}
Coupled stochastic derivatives differentiate programs with discrete random choices \citep{arya2022automatic}, and \citet{arya2023differentiating,arya2024gradient} specialize them to MH by coupling accepted and rejected continuations. We apply this estimator to a terminal discriminator reward, subsample decisions with inverse inclusion probabilities \citep{horvitz1952generalization}, and compare it with a score-function estimator in Section~\ref{sec:exact}.

\paragraph{Graph generation and reward adaptation.}
Discrete diffusion and flow models generate graphs by learned denoising \citep{vignac2023digress,campbell2022continuous,siraudin2024cometh}, whereas GEM samples from a learned potential with MH refinement. Reward fine-tuning methods such as GDPO, DRAKES, and Graph-GRPO update the denoiser or flow network \citep{liu2024gdpo,wang2025drakes,zhu2026graphgrpo}, and TFG-Flow and clean-sample Markov chains apply a reward or conditioning signal during sampling \citep{lin2025tfg,phunyaphibarn2026clean}. Here the pretrained networks stay fixed; only the acceptance energy changes, contrasting a model with a weaker version of itself as in autoguidance \citep{karras2024guiding}.

\section{Finite-Horizon Energy Learning}
\label{sec:method}

\subsection{Sampler and Terminal Law}
\label{sec:objective}
Let $p$ be the data distribution on a finite discrete state space $\mathcal X$; for graphs, a state assigns a categorical label to every node and node pair, including a no-edge category. Starting from $X_0\sim q_0$, at steps $t=0,\ldots,K-1$ the sampler proposes $Y_t\sim Q_t(\cdot\mid X_t)$ and accepts it with the MH probability
\begin{equation}
\alpha_{\theta,t}(x,y)=\min\!\left\{1,
e^{-\beta_t[E_\theta(y)-E_\theta(x)]}
\frac{Q_t(x\mid y)}{Q_t(y\mid x)}\right\};
\label{eq:acceptance}
\end{equation}
rejection leaves the state unchanged. Writing $P_{\theta,t}$ for the resulting transition kernel, the terminal law after $K$ steps is
\begin{equation}
q_{\theta,K}=q_0P_{\theta,0}P_{\theta,1}\cdots P_{\theta,K-1}.
\label{eq:finite-law}
\end{equation}
The initialization $q_0$, the proposal kernel $Q_t$, and the inverse-temperature schedule $\beta_t$ do not depend on $\theta$, which enters only through the acceptance energy $E_\theta$. A fixed proposal kernel does not imply identical realized proposals: once two runs make different decisions, later proposals start from different states. The annealed kernels need not share a common invariant distribution; our objective is defined on the terminal law $q_{\theta,K}$, not on the equilibrium density $\propto e^{-\beta E_\theta}$ at a fixed temperature.

\subsection{Fixed-Discriminator Objective}
\label{sec:discriminator-objective}
A discriminator with logit $d_\psi(x)$ is trained to distinguish data from samples of $q_{\theta,K}$. With $\sigma$ the logistic function and terminal reward $r_\psi(x)=\log\sigma(d_\psi(x))$, each energy update ascends the non-saturating generator objective \citep{goodfellow2014generative}
\begin{equation}
J(\theta;\psi)=\E_{X_K\sim q_{\theta,K}}[r_\psi(X_K)]
\label{eq:objective}
\end{equation}
with $\psi$ held fixed, and the discriminator is refitted between updates. If the discriminator is the optimal unrestricted classifier for the balanced, unweighted population problem, held fixed at the current generator, and $p>0$ wherever $q_{\theta,K}>0$, then $-\nabla_\theta J$ is the gradient of a divergence between $q_{\theta,K}$ and $p$ (Appendix~\ref{app:exact}), which Section~\ref{sec:exact} uses as a numerical reference. The weighted, finitely trained molecular discriminator need not satisfy these conditions; the estimator below holds for any fixed discriminator.

Exact likelihood training of an energy takes its negative-phase expectation under the equilibrium model distribution. GEM's contrastive term instead raises energy on generated negatives held fixed during differentiation \citep[Section~3.4 and Appendix~I]{balcerak2026gem}; with short-chain negatives, such a detached update need not equal the exact likelihood gradient \citep{hinton2002training,agoritsas2023explaining}. Even exact likelihood training targets the equilibrium law: with population data and an energy family that contains the target, its optimum is the equilibrium-correct energy, which need not be the best acceptance energy for a $K$-step run. Equation~\eqref{eq:objective} instead scores the returned samples.

\subsection{Gradient Estimation with Paired Continuations}
\label{sec:estimator}
\paragraph{Continuation values.}
Let $V_{t+1}(s)=\E[r_\psi(X_K)\mid X_{t+1}=s]$ be the expected terminal reward of a chain that is in state $s$ after step $t$ and continues under the current energy and the remaining schedule. Because $q_0$ and $Q_t$ do not depend on $\theta$, the product rule applied to \eqref{eq:finite-law} gives \citep{arya2023differentiating,arya2024gradient}
\begin{equation}
\nabla_\theta J=\sum_{t=0}^{K-1}\E\bigl[\Delta V_{t+1}\,\nabla_\theta\alpha_{\theta,t}(X_t,Y_t)\bigr],
\label{eq:identity}
\end{equation}
where $\Delta V_{t+1}=V_{t+1}(Y_t)-V_{t+1}(X_t)$ and $(X_t,Y_t)$ are the state and proposal of a chain run with the current energy (Appendix~\ref{app:gradient}). Each term differentiates only the local acceptance probability; the effect of the decision on later steps enters through $\Delta V_{t+1}$, whose sign indicates whether accepting or rejecting yields the higher expected terminal reward.

\paragraph{Paired continuations.}
The two possible next states are $X_{t+1}^{+}=Y_t$ and $X_{t+1}^{-}=X_t$. Continuing each to step $K$ gives rewards $R^\pm=r_\psi(X_K^\pm)$, and $\E[R^+-R^-\mid X_t,Y_t]=\Delta V_{t+1}$ whenever each branch follows the correct transition law. Both continuations use the same future random numbers, mapped through their own proposal distributions; this coupling changes only the joint law of the branches and can reduce the variance of $R^+-R^-$. One branch coincides with the base trajectory, so only the other is simulated. Branches that reach the same state with the same carried sampler quantities agree from then on and can be merged.

\paragraph{Subsampled decisions.}
To limit simulation, we select a subset $\mathcal S$ of decisions from $N$ base chains, including step $t$ of chain $i$ with probability $p_{it}$, and use
\begin{equation}
\widehat g=\frac1N\sum_{(i,t)\in\mathcal S}
\frac{R^+_{it}-R^-_{it}}{p_{it}}\,
\nabla_\theta\alpha_{\theta,t}(X_{it},Y_{it}).
\label{eq:estimator}
\end{equation}
For a fixed discriminator, $\widehat g$ is unbiased for \eqref{eq:identity} if (i) the acceptance probabilities are differentiable at the sampled decisions, (ii) both continuations follow the correct law, (iii) $p_{it}>0$ wherever $\nabla_\theta\alpha_{\theta,t}\neq0$, and (iv) selection depends only on the chain prefix, the proposal, and quantities fixed before the update, not on continuation randomness. The sum is normalized by $N$, not by $|\mathcal S|$, and states, rewards, and selection probabilities are held fixed during differentiation. At unsaturated proposals, $\nabla_\theta\alpha_{\theta,t}=-\beta_t\alpha_{\theta,t}\nabla_\theta[E_\theta(Y_t)-E_\theta(X_t)]$, and strictly saturated proposals have zero derivative. At an MH tie, where the acceptance ratio equals one while its derivative is nonzero, $\alpha_{\theta,t}$ is not differentiable; the implementation uses zero there, and the unbiasedness statement excludes such points. The statement also does not extend to clipping, coordinate constraints, adaptive optimization, discriminator refitting, or convergence of the alternating procedure (Appendix~\ref{app:gradient}).

\begin{algorithm}[t]
\caption{Finite-horizon energy learning}
\label{alg:learning}
\begin{algorithmic}[1]
\Require Data $\mathcal D$; initial $\theta,\psi$; fixed $q_0,Q_t,\beta_t$; horizon $K$; $N$ base chains
\For{each energy update}
\State Freeze $\theta$, $\psi$, and the decision-selection rule
\State Simulate $N$ chains to $K$; cache random draws
\State Select decisions $\mathcal S$ with probabilities $p_{it}$
\For{$(i,t)\in\mathcal S$}
\State $X^+_{i,t+1}\gets Y_{it},\quad X^-_{i,t+1}\gets X_{it}$
\State Continue $X^+,X^-$ to $K$ with shared draws
\State $R^\pm_{it}\gets r_\psi(X^\pm_{iK})$
\EndFor
\State Compute $\widehat g$ by \eqref{eq:estimator}; $\theta\gets\operatorname{Update}(\theta,\widehat g)$
\State Refit $d_\psi$ on $\mathcal D$ and $\{X_{iK}\}_{i=1}^N$
\State Refresh the selection rule for the next update
\EndFor
\State \Return $E_\theta$ \Comment{used in a single trajectory}
\end{algorithmic}
\end{algorithm}

\subsection{Training and Generation}
\label{sec:training}
Algorithm~\ref{alg:learning} starts from an initial energy and a discriminator fitted to data and initial generated samples, and alternates one ascent step $\operatorname{Update}(\theta,\widehat g)$ with discriminator refitting. The exact-state experiments select one uniformly random decision per chain; molecular training selects about four per chain, with inclusion probabilities increasing with the acceptance-gradient magnitude, scaled using the preceding update (Appendix~\ref{app:implementation}). Both studies report the final iterate, without test-based checkpoint selection; Appendix~\ref{app:resources} bounds the transitions simulated per update. At generation time, the learned energy replaces the original acceptance energy in the same $K$-step sampler, without the discriminator or continuations; the per-step cost can still depend on the energy (Section~\ref{sec:molecules}).

\FloatBarrier
\section{Exact Finite-State Study}
\label{sec:exact}

We first ask whether an energy that is exact at equilibrium can be improved for a short run, and whether a scalar temperature change explains the improvement. Small state spaces allow exact terminal laws, while learning uses sampled trajectories.

\begin{figure*}[t]
\centering
\includegraphics[width=\textwidth]{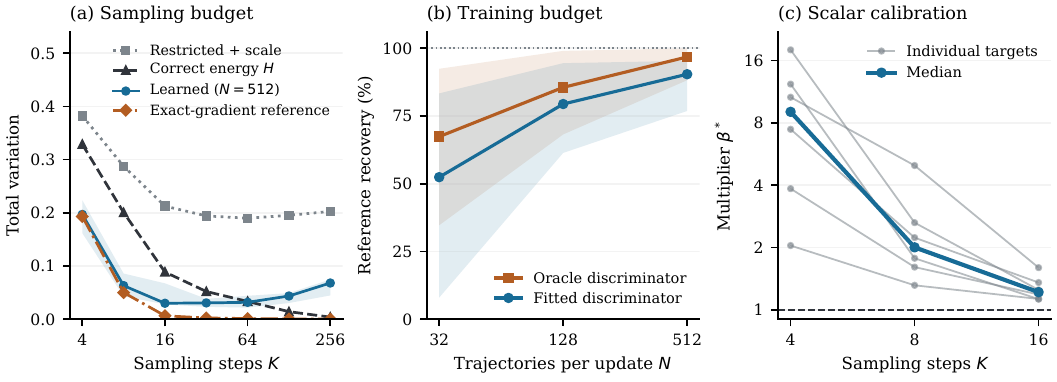}
\caption{\textbf{Finite-time sampling on the reference targets (original runs).} (a) Exact terminal total variation (TV) against the number of steps $K$; each learned point is an energy trained separately for that horizon with $N=512$ trajectories per update. Curves are medians over six targets after aggregating training seeds. (b) Recovery (\%) of the numerical reference's improvement over the restricted energy, for learning with a fitted or an oracle discriminator, against trajectories per update $N$; 100\% is not an upper bound. (c) TV-minimizing multiplier $\beta^*$ of $E=\beta H$ under the shared annealing schedule, for individual targets (thin) and their median (thick); the dashed line marks $\beta=1$. Shading shows interquartile ranges, not confidence intervals. Horizontal axes are logarithmic, as is the vertical axis in (c).}
\label{fig:exact-learning}
\end{figure*}

\subsection{Targets, Sampler, and Comparisons}
\label{sec:exact-setup}
Targets are synthetic pairwise binary models $p\propto e^{-H}$ on $L=6$ or $L=8$ spins (64 or 256 configurations) with random fields and couplings (Appendix~\ref{app:exact}). For each $L\in\{6,8\}$, the \emph{reference} set contains three independently drawn target distributions, and the \emph{new-target} set contains three further draws. Each target is specified by a fixed set of unary fields and pairwise couplings, held constant across horizons and training seeds. We use identifiers such as \texttt{Ref-6-1} and \texttt{New-8-3} to specify the target set, number of spins, and instance index. The index identifies a distribution, not a training seed. Training is repeated from scratch on each new target; no fitted energy is transferred between targets. The sampler starts from the product of the target's one-variable marginals, proposes a flip of one uniformly chosen spin, and raises the inverse temperature from 0.2 to 1 over the first half of the run before holding it fixed. Every compared energy uses this sampler at a given horizon; unlike GEM's fixed-length ramp, the schedule scales with $K$. We compare:
\begin{itemize}
\item \emph{Restricted energy:} unary and ring interactions, fitted by population maximum likelihood, scalar-calibrated for TV; learning starts from it.
\item \emph{Correct $H$:} the target energy, recovered by population pseudo-likelihood \citep{besag1975pseudolikelihood}, exact at equilibrium.
\item \emph{Scalar-tuned $\beta H$:} the multiple of $H$ that minimizes exact terminal TV, an oracle control.
\item \emph{Learned, fitted discriminator (ours):} all unary and pairwise coefficients (21 for $L=6$, 36 for $L=8$), a family that contains $H$, trained by Algorithm~\ref{alg:learning} with a discriminator linear in the same features.
\item \emph{Learned, oracle discriminator:} the same sampled training, with rewards from the exact density ratio between the target and the current generator.
\item \emph{Numerical reference:} exact-gradient minimization over the learned family of the divergence $D(q\|p)=2\KL(\tfrac12(p+q)\,\|\,p)$ of Section~\ref{sec:discriminator-objective}; it is not optimized for TV and not certified optimal.
\end{itemize}
Learned energies use 2,000 updates, $N=512$ trajectories per update unless stated otherwise, three training seeds, and the final iterate. The fitted discriminator starts from the exact population logistic fit and is then refitted on target and generated samples; exact marginals and population fits are idealizations of this study. We report total variation, $\TV(q,p)=\tfrac12\sum_x|q(x)-p(x)|$, computed exactly, so reported variation comes only from targets and training seeds.

\subsection{Comparison with the Equilibrium-Correct Energy}
At $K=4,8,16$, the learned energy has lower median TV than $H$ in all 18 target--horizon settings of the reference set (Figure~\ref{fig:exact-learning}a, Table~\ref{tab:exact-main}). A fresh-seed replication on the same targets again improves all 18 settings, with 53 of 54 individual fits better than $H$. On the new targets, learning improves all 18 settings and all 54 fits; median TV falls from 0.553 to 0.518, from 0.413 to 0.307, and from 0.240 to 0.069 at $K=4,8,16$. Because $H$ is exact at equilibrium, these gains correct the finite-run law rather than the density. Horizons within a target are dependent, so the counts describe the experiment rather than constitute independent tests.

\begin{table}[t]
\centering
\caption{\textbf{Exact terminal TV at short horizons} (lower is better). Entries are medians over six targets of each target's median over three training seeds. Reference rows summarize the original runs in Figure~\ref{fig:exact-learning}a; Table~\ref{tab:exact-detail} reports per-target results for a fresh-seed replication on the reference targets and for the new targets. Bold marks the smaller median.}
\label{tab:exact-main}
\begin{tabular}{@{}lrcc@{}}
\toprule
Targets & $K$ & Correct $H$ & Learned (ours) \\
\midrule
Reference & 4 & $0.329$ & $\mathbf{0.197}$ \\
 (original runs) & 8 & $0.201$ & $\mathbf{0.063}$ \\
 & 16 & $0.089$ & $\mathbf{0.030}$ \\
\midrule
New targets & 4 & $0.553$ & $\mathbf{0.518}$ \\
 & 8 & $0.413$ & $\mathbf{0.307}$ \\
 & 16 & $0.240$ & $\mathbf{0.069}$ \\
\bottomrule
\end{tabular}

\end{table}

\subsection{Comparison with Scalar Tempering}
Scalar tempering tests whether rescaling $H$ explains these gains. Learned energies have lower TV than scalar-tuned $\beta H$ in 14 of 18 reference settings, in both the original runs and the replication, and in 15 of 18 new-target settings. The three new-target exceptions all occur on \texttt{New-8-2}, where the scalar-tuned control has lower TV at each tested horizon (Table~\ref{tab:exact-detail}). The control is an oracle, since it uses the target and exact terminal probabilities. Its optimal multiplier is far from one at short horizons: the median over reference targets falls from 9.04 at $K=4$ to 2.00 at $K=8$ and 1.22 at $K=16$ (Figure~\ref{fig:exact-learning}c). This is a target-specific calibration under the horizon-scaled schedule, not a general law relating temperature to budget.

\subsection{Training Budget, Estimators, and Long Horizons}
Figure~\ref{fig:exact-learning}b varies the number of training trajectories per update, $N$. For each target and horizon, with TV errors $e_{\rm base}$, $e_{\rm learned}$, and $e_{\rm ref}$ of the restricted, learned, and reference energies, the recovery is $100(e_{\rm base}-e_{\rm learned})/(e_{\rm base}-e_{\rm ref})$, computed in each of the 42 settings (six targets, seven horizons) before taking medians. Zero means no gain over the restricted energy and 100 means matching the reference; because the restricted family is misspecified, the gain combines density and finite-time corrections. With the fitted discriminator, median recovery rises from 52.4\% at $N=32$ to 79.4\% at 128 and 90.4\% at 512, where the median TV gap to the reference is 0.024. With the oracle discriminator, it reaches 96.8\% at $N=512$. At matched numbers of simulated transitions, a leave-one-out score-function estimator also improves on $H$ in all 18 short-horizon reference settings; coupling has lower median TV in 12 of them (median difference $-0.0015$) but took 4.66 times the CPU time (Appendix~\ref{app:controls}). Terminal feedback is thus effective with either estimator, without an efficiency advantage for coupling.

Learning is least effective at long horizons. Each point in Figure~\ref{fig:exact-learning}a is a separately trained energy; beyond $K=64$, the learned energies are less accurate than $H$, while the numerical reference stays near zero TV. Because the learned family contains both $H$ and the reference solution, the shortfall arises in sampled training, not in the energy family. Discriminator accuracy, gradient variance over long continuations, and the fixed optimization budget are candidate causes that this study does not separate.

\section{Adapting Pretrained GEM Potentials on MOSES}
\label{sec:molecules}

We next adapt a pretrained graph sampler, changing only the acceptance energy and comparing at matched numbers of refinement steps.

\begin{table*}[t]
\centering
\renewcommand{\arraystretch}{1.12}
\caption{\textbf{GEM with and without finite-horizon energy learning} at matched refinement steps $K$, with exact MH acceptance. Mean $\pm$ sample SD over three paired seeds, each with 25,000 requested graphs; validity and VUN are \% of requested graphs. Bold marks the better mean in each pair, not significance.}
\label{tab:molecular}
\begin{tabular*}{.98\textwidth}{@{\extracolsep{\fill}}rlrrr@{}}
\toprule
$K$ & Method & FCD $\downarrow$ & Validity $\uparrow$ & VUN $\uparrow$ \\
\midrule
125 & GEM (calibrated) & $2.508\pm0.026$ & $87.13\pm0.14$ & $85.15\pm0.09$ \\
 & GEM + learned energy (Ours) & $\mathbf{2.366\pm0.063}$ & $\mathbf{92.55\pm0.24}$ & $\mathbf{90.44\pm0.25}$ \\
\midrule
250 & GEM (calibrated) & $2.044\pm0.030$ & $\mathbf{95.43\pm0.10}$ & $\mathbf{92.76\pm0.05}$ \\
 & GEM + learned energy (Ours) & $\mathbf{1.829\pm0.021}$ & $94.65\pm0.33$ & $92.45\pm0.29$ \\
\midrule
1000 & GEM (calibrated) & $1.561\pm0.018$ & $\mathbf{97.84\pm0.14}$ & $93.77\pm0.12$ \\
 & GEM + learned energy (Ours) & $\mathbf{1.504\pm0.018}$ & $97.82\pm0.03$ & $\mathbf{93.92\pm0.09}$ \\
\bottomrule
\end{tabular*}

\end{table*}

\subsection{Two-Coefficient Energy and Controlled Comparison}
\label{sec:molecules-setup}
We use the pretrained GEM model \citep{balcerak2026gem} on MOSES \citep{polykovskiy2020moses}. Each chain fixes a node count drawn from the training distribution. GEM initializes a random graph with at most 225 greedy edits under its released potential $\VG$; $K$ refinement steps then use gradient-informed categorical proposals and the released inverse-temperature schedule, which rises from about 0.78 to 17.62 over 500 steps and then stays constant. We retain this initialization, proposal mechanism, and schedule. $K$ counts refinement steps after initialization, not all edits, network evaluations, or elapsed computation, and runs with $K<500$ stop during the ramp. The learned acceptance energy is
\begin{equation}
E_\theta(x)=c\bigl[\VG(x)+\omega\{\VG(x)-\Vpre(x)\}\bigr],
\label{eq:energy}
\end{equation}
where $\Vpre$ is GEM's frozen transport-only predecessor checkpoint, $c$ sets the energy scale, and $\omega$ weights the difference between checkpoints, as in autoguidance \citep{karras2024guiding}. Equation~\eqref{eq:energy} is the learned energy $E_\theta$ of Figure~\ref{fig:method}, with $\theta=(c,\omega)$; both potentials stay frozen, and the combination remains invariant to node relabeling. Rewards come from a training-only discriminator, a LoRA-adapted copy of $\VG$ with about 9.7 million trainable parameters \citep{hu2022lora}.

The baseline at each budget is GEM with exact MH acceptance and an acceptance scale calibrated on development FCD: 0.70 at $K=125,250$ and 0.865 at $K=1000$, with $\omega=0$. Learning starts from this calibrated energy. Exact MH differs from GEM's released acceptance rule in how proposal retry stages enter the acceptance ratio (Appendix~\ref{app:molecular}); the released rule is used only to reuse the trained energies without retraining (Table~\ref{tab:released}). Training uses 80 energy updates with 256 base chains and about four selected decisions per chain. The discriminator is fitted to generated graphs and node-count-matched training molecules, with invalid generated graphs up-weighted, and is refitted after each update (Appendix~\ref{app:implementation}). The energy-update objective uses discriminator rewards rather than FCD; development FCD informs baseline calibration and discriminator screening, and it selects the static-coefficient controls in Section~\ref{sec:molecular-scope}.

\subsection{Metrics and Decision Rule}
We report Fr\'echet ChemNet distance (FCD) \citep{preuer2018fcd}, validity, and VUN. FCD compares the means and covariances of ChemNet features of generated molecules with those of the MOSES scaffold-test reference of 176,225 molecules, using the canonical largest fragment of each valid output and retaining duplicates. Validity is the percentage of requested graphs that decode and sanitize; following GEM's convention, a disconnected but sanitizable molecule counts as valid. VUN is the percentage of requested graphs that are valid, unique, and absent from the training set. Failed outputs are retained, so both rates use all requested graphs as denominators. Development FCD instead uses a fixed reference of 16,010 molecules selected deterministically from the MOSES training data, with 2,048 generated graphs per candidate from a generation stream separate from the final evaluation; this split does not imply that these molecules were excluded from the pretraining data of the released GEM checkpoints. Because encoding and processing choices affect the scores, they are compared within this implementation, not with published leaderboard values.

Each comparison uses three seeds with 25,000 requested graphs each; both methods share the generation seed paired with each training seed, as well as the initialization, proposals, schedule, decoder, and evaluator, and seed-level differences are the units of replication. An internally prespecified joint rule requires lower FCD in a one-sided paired $t$ test at level 0.05 and one-sided 95\% lower confidence bounds above $-1$ percentage point for the validity and VUN differences. The exact-MH comparison at $K=250$ was the original primary comparison; the other budgets and the released-rule analysis followed it, and no multiplicity adjustment is applied (Appendix~\ref{app:molecular-protocol}). Matched steps are not matched compute: at $K=250$, generation took 0.932 H100 GPU-seconds per requested graph with the learned energy versus 0.859 for calibrated GEM.

\subsection{Results}
\begin{figure}[t]
\centering
\includegraphics[width=\linewidth]{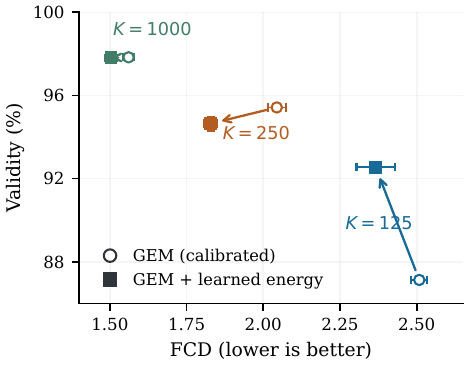}
\caption{\textbf{FCD--validity changes at matched budgets.} Circles: calibrated GEM; squares: GEM with the learned energy (both exact MH); arrows join matched-budget pairs. FCD decreases at every budget, whereas validity rises at $K=125$, falls at $K=250$, and is nearly unchanged at $K=1000$. Bars show one sample SD over three seeds; each budget has its own calibrated baseline and trained coefficients.}
\label{fig:molecular-quality}
\end{figure}

The learned energy lowers FCD in every paired seed at all three budgets (Table~\ref{tab:molecular}, Figure~\ref{fig:molecular-quality}); relative changes use unrounded seed means. At $K=125$, all three metrics improve: FCD decreases by 5.7\% (2.508 to 2.366), validity rises by 5.42 points, and VUN by 5.29 points; these yield gains are relative to a baseline scale selected for FCD and partly reflect that choice. At $K=250$, FCD decreases by 10.5\% (2.044 to 1.829), while mean validity and VUN decrease by 0.78 and 0.32 points. The mean validity decrease was below one percentage point, but its one-sided lower confidence bound ($-1.25$ points) crossed the prespecified margin; non-inferiority was therefore not established, and this original primary comparison does not meet the joint criterion. At $K=1000$, FCD decreases by 3.7\% (1.561 to 1.504), validity and VUN change by $-0.02$ and $+0.15$ points. The joint criterion is met at $K=125$ and $1000$.

The $K=125$ yield gains persist under strict scoring that requires connected molecules and passes stricter representation checks (validity $+4.77$, VUN $+4.65$ points), and equalizing the number of valid molecules entering FCD changes the paired difference only from $-0.142$ to $-0.138$. These checks reuse the generated samples and address evaluation sensitivity, not the effect of encoding mismatches in the discriminator's training pool. Reusing the learned energies without retraining under GEM's released acceptance rule again lowers FCD at all budgets. The joint criterion holds at $K=250$ and $1000$ but not at $K=125$, which used two prespecified seeds and an uncalibrated released-GEM baseline (Table~\ref{tab:released}); these reuse analyses are not independent replications.

\subsection{Same-Family Controls and Protocol Conditions}
\label{sec:molecular-scope}
The baseline calibrates one coefficient while learning adjusts two, so Table~\ref{tab:molecular} does not separate the effect of gradient-based learning from that of a second coefficient. Static two-coefficient settings selected on development FCD reach FCD 2.280 at $K=125$ (learned: 2.366), with lower validity (91.10\% versus 92.55\%); this FCD difference is unresolved with three seeds. At $K=250$, the static setting reaches 1.811 (learned: 1.829) but fails its development validity threshold. Finite-difference training under a different optimization schedule reaches exact-rule FCD 1.821. These controls use different selection and optimization budgets and compute costs, and there is no corresponding three-seed static evaluation at $K=1000$ (Appendix~\ref{app:molecular-controls}). The reported comparisons do not establish an FCD advantage over coefficient selection or finite-difference training within the same energy family. Lower FCD also does not imply that every molecular property improves: molecular-weight distance worsens at $K=1000$.

Three protocol conditions affect interpretation (Appendix~\ref{app:molecular-protocol}). At $K=125$, the allowed rate of re-encoding mismatches in the discriminator's training pool was raised from 1\% to 2\% before the pool was generated; the pool had 1.65\% mismatches, so the original threshold would have stopped the experiment. At $K=1000$, the discriminator passed its development screen only with a revised negative pool. At $K=500$, the screen failed and no energy was trained, so that budget is missing, not an evaluated failure.

\balance
\section{Discussion and Conclusion}
\label{sec:discussion}

We formulated finite-horizon energy learning to optimize the distribution returned by an MH sampler while retaining its initialization, proposal kernel, and annealing schedule. The finite-state experiments isolate a benefit of this objective: learned energies improve median terminal TV over an equilibrium-correct energy in every tested short-horizon target--horizon setting, and over its numerically tuned scalar multiple in most settings. These gains concern the finite-run distribution rather than correction of an inaccurate equilibrium density.

On MOSES, adapting two coefficients of frozen GEM potentials reduces FCD by 3.7--10.5\% relative to scale-calibrated GEM with exact MH acceptance at matched refinement steps. The adapted generator uses a single trajectory without discriminator evaluations or additional continuations. The molecular results do not establish superiority over same-family coefficient selection, and the original $K=250$ comparison does not establish validity non-inferiority. Thus, the observed FCD gains should not be interpreted as uniformly improved sample quality or reduced computational cost.

The present procedure trains a separate energy for each horizon, and sampled training falls behind the correct energy at long horizons in the finite-state study. Training across budgets and understanding this long-horizon shortfall remain directions for further work. The results support acceptance-energy learning as a means of adapting finite-run generation while retaining the proposal mechanism.

\label{main:end}
\clearpage
\label{main:after}
\nobalance
\subsection*{AI use statement}
Generative AI tools were used to assist with drafting portions of the manuscript and revising the text for clarity, organization, consistency checks and readability. The authors take full responsibility for the final manuscript, including its originality, technical claims, reported results, and references.

\bibliography{references}

\begin{thebibliography}{}

\bibitem[Agoritsas et~al., 2023]{agoritsas2023explaining}
Agoritsas, E., Catania, G., Decelle, A., and Seoane, B. (2023).
\newblock Explaining the effects of non-convergent {MCMC} in the training of energy-based models.
\newblock In {\em Proceedings of the 40th International Conference on Machine Learning}, volume 202 of {\em Proceedings of Machine Learning Research}.

\bibitem[Arya et~al., 2022]{arya2022automatic}
Arya, G., Schauer, M., Sch{\"a}fer, F., and Rackauckas, C. (2022).
\newblock Automatic differentiation of programs with discrete randomness.
\newblock In {\em Advances in Neural Information Processing Systems}, volume~35.

\bibitem[Arya et~al., 2024]{arya2024gradient}
Arya, G., Schauer, M., and Seyer, R. (2024).
\newblock Gradient estimation via differentiable {M}etropolis-{H}astings.
\newblock {\em arXiv preprint arXiv:2406.14451}.

\bibitem[Arya et~al., 2023]{arya2023differentiating}
Arya, G., Seyer, R., Sch{\"a}fer, F., Chandra, K., Lew, A.~K., Huot, M., Mansinghka, V.~K., Ragan-Kelley, J., Rackauckas, C., and Schauer, M. (2023).
\newblock Differentiating {M}etropolis-{H}astings to optimize intractable densities.
\newblock In {\em ICML 2023 Workshop on Differentiable Almost Everything}.
\newblock arXiv:2306.07961.

\bibitem[Balcerak et~al., 2026]{balcerak2026gem}
Balcerak, M., Shit, S., Prabhakar, C., Kaltenbach, S., Albergo, M.~S., Du, Y., and Menze, B. (2026).
\newblock Graph energy matching: Transport-aligned energy-based modeling for graph generation.
\newblock {\em arXiv preprint arXiv:2603.23398}.

\bibitem[Besag, 1975]{besag1975pseudolikelihood}
Besag, J. (1975).
\newblock Statistical analysis of non-lattice data.
\newblock {\em Journal of the Royal Statistical Society. Series D (The Statistician)}, 24(3):179--195.

\bibitem[Campbell et~al., 2022]{campbell2022continuous}
Campbell, A., Benton, J., De~Bortoli, V., Rainforth, T., Deligiannidis, G., and Doucet, A. (2022).
\newblock {A Continuous Time Framework for Discrete Denoising Models}.
\newblock In {\em Advances in Neural Information Processing Systems}.

\bibitem[Dai et~al., 2020]{dai2020aloe}
Dai, H., Singh, R., Dai, B., Sutton, C., and Schuurmans, D. (2020).
\newblock Learning discrete energy-based models via auxiliary-variable local exploration.
\newblock In {\em Advances in Neural Information Processing Systems}, volume~33.

\bibitem[Goodfellow et~al., 2014]{goodfellow2014generative}
Goodfellow, I., Pouget-Abadie, J., Mirza, M., Xu, B., Warde-Farley, D., Ozair, S., Courville, A., and Bengio, Y. (2014).
\newblock Generative adversarial nets.
\newblock In {\em Advances in Neural Information Processing Systems}, volume~27.

\bibitem[Hinton, 2002]{hinton2002training}
Hinton, G.~E. (2002).
\newblock Training products of experts by minimizing contrastive divergence.
\newblock {\em Neural Computation}, 14(8):1771--1800.

\bibitem[Horvitz and Thompson, 1952]{horvitz1952generalization}
Horvitz, D.~G. and Thompson, D.~J. (1952).
\newblock A generalization of sampling without replacement from a finite universe.
\newblock {\em Journal of the American Statistical Association}, 47:663--685.

\bibitem[Hu et~al., 2022]{hu2022lora}
Hu, E.~J., Shen, Y., Wallis, P., Allen-Zhu, Z., Li, Y., Wang, S., Wang, L., and Chen, W. (2022).
\newblock {LoRA}: Low-rank adaptation of large language models.
\newblock In {\em International Conference on Learning Representations}.

\bibitem[Karras et~al., 2024]{karras2024guiding}
Karras, T., Aittala, M., Kynk{\"a}{\"a}nniemi, T., Lehtinen, J., Aila, T., and Laine, S. (2024).
\newblock Guiding a diffusion model with a bad version of itself.
\newblock In {\em Advances in Neural Information Processing Systems}, volume~37.

\bibitem[Lin et~al., 2025]{lin2025tfg}
Lin, H., Li, S., Ye, H., Yang, Y., Ermon, S., Liang, Y., and Ma, J. (2025).
\newblock {TFG-Flow: Training-free Guidance in Multimodal Generative Flow}.
\newblock In {\em International Conference on Learning Representations}.

\bibitem[Liu et~al., 2024]{liu2024gdpo}
Liu, Y., Du, C., Pang, T., Li, C., Lin, M., and Chen, W. (2024).
\newblock {Graph Diffusion Policy Optimization}.
\newblock In {\em Advances in Neural Information Processing Systems}.

\bibitem[Nijkamp et~al., 2019]{nijkamp2019learning}
Nijkamp, E., Hill, M., Zhu, S.-C., and Wu, Y.~N. (2019).
\newblock Learning non-convergent non-persistent short-run {MCMC} toward energy-based model.
\newblock In {\em Advances in Neural Information Processing Systems}, volume~32.

\bibitem[Phunyaphibarn and Sung, 2026]{phunyaphibarn2026clean}
Phunyaphibarn, P. and Sung, M. (2026).
\newblock {Reward-Guided Discrete Diffusion via Clean-Sample Markov Chain for Molecule and Biological Sequence Design}.
\newblock {\em arXiv preprint arXiv:2602.09424}.

\bibitem[Polykovskiy et~al., 2020]{polykovskiy2020moses}
Polykovskiy, D., Zhebrak, A., Sanchez-Lengeling, B., Golovanov, S., Tatanov, O., Belyaev, S., Kurbanov, R., Artamonov, A., Aladinskiy, V., Veselov, M., Kadurin, A., Johansson, S., Chen, H., Nikolenko, S., Aspuru-Guzik, A., and Zhavoronkov, A. (2020).
\newblock Molecular sets ({MOSES}): A benchmarking platform for molecular generation models.
\newblock {\em Frontiers in Pharmacology}, 11:565644.

\bibitem[Preuer et~al., 2018]{preuer2018fcd}
Preuer, K., Renz, P., Unterthiner, T., Hochreiter, S., and Klambauer, G. (2018).
\newblock Fr{\'e}chet {ChemNet} distance: A metric for generative models for molecules in drug discovery.
\newblock {\em Journal of Chemical Information and Modeling}, 58(9):1736--1741.

\bibitem[Qin et~al., 2025]{qin2025defog}
Qin, Y., Madeira, M., Thanou, D., and Frossard, P. (2025).
\newblock {DeFoG: Discrete Flow Matching for Graph Generation}.
\newblock In {\em Proceedings of the 42nd International Conference on Machine Learning}.

\bibitem[Siraudin et~al., 2024]{siraudin2024cometh}
Siraudin, A., Malliaros, F.~D., and Morris, C. (2024).
\newblock {COMETH: A Continuous-Time Discrete-State Graph Diffusion Model}.
\newblock {\em arXiv preprint arXiv:2406.06449}.

\bibitem[Song et~al., 2017]{song2017anicemc}
Song, J., Zhao, S., and Ermon, S. (2017).
\newblock {A-NICE-MC}: Adversarial training for {MCMC}.
\newblock In {\em Advances in Neural Information Processing Systems}, volume~30.

\bibitem[Vignac et~al., 2023]{vignac2023digress}
Vignac, C., Krawczuk, I., Siraudin, A., Wang, B., Cevher, V., and Frossard, P. (2023).
\newblock {DiGress: Discrete Denoising Diffusion for Graph Generation}.
\newblock In {\em International Conference on Learning Representations}.

\bibitem[Wang et~al., 2025a]{wang2025reinforcement}
Wang, C., Chen, W.~Y., Kanagawa, H., and Oates, C.~J. (2025a).
\newblock Reinforcement learning for adaptive {MCMC}.
\newblock In {\em Proceedings of The 28th International Conference on Artificial Intelligence and Statistics}, volume 258 of {\em Proceedings of Machine Learning Research}, pages 640--648.

\bibitem[Wang et~al., 2025b]{wang2025drakes}
Wang, C., Uehara, M., He, Y., Wang, A., Biancalani, T., Lal, A., Jaakkola, T., Levine, S., Wang, H., and Regev, A. (2025b).
\newblock {Fine-Tuning Discrete Diffusion Models via Reward Optimization with Applications to DNA and Protein Design}.
\newblock In {\em International Conference on Learning Representations}.

\bibitem[Zhang et~al., 2022a]{zhang2022ebgfn}
Zhang, D., Malkin, N., Liu, Z., Volokhova, A., Courville, A., and Bengio, Y. (2022a).
\newblock Generative flow networks for discrete probabilistic modeling.
\newblock In {\em Proceedings of the 39th International Conference on Machine Learning}, volume 162 of {\em Proceedings of Machine Learning Research}, pages 26412--26428.

\bibitem[Zhang et~al., 2022b]{zhang2022langevin}
Zhang, R., Liu, X., and Liu, Q. (2022b).
\newblock {A Langevin-like Sampler for Discrete Distributions}.
\newblock In {\em Proceedings of the 39th International Conference on Machine Learning}.

\bibitem[Zhu et~al., 2026]{zhu2026graphgrpo}
Zhu, B., Bo, D., Zhang, D.~C., and Wang, X. (2026).
\newblock {Graph-GRPO: Training Graph Flow Models with Reinforcement Learning}.
\newblock {\em arXiv preprint arXiv:2603.10395}.

\end{thebibliography}

\label{paper:end}
\clearpage
\appendix
\onecolumn
\aistatstitle{\papertitle:\\ Supplementary Material}

\noindent Appendix~\ref{app:implementation} derives the gradient estimator and describes the implementation, Appendix~\ref{app:exact} describes the finite-state experiments, and Appendix~\ref{app:molecular} describes the molecular evaluation.
\section{Gradient Derivation and Implementation}
\label{app:implementation}

\subsection{Gradient Identity and Estimator}
\label{app:gradient}
We specialize the coupled MH derivative of \citet{arya2023differentiating,arya2024gradient} to the finite-horizon objective. Fix the discriminator, write $r=r_\psi$ as a column vector over $\mathcal X$, and abbreviate $P_t=P_{\theta,t}$ and $\alpha_t=\alpha_{\theta,t}$, so that $J=q_0P_0\cdots P_{K-1}r$. Because $q_0$ does not depend on $\theta$, the product rule gives
\[
\nabla_\theta J=\sum_{t=0}^{K-1}q_t\,(\nabla_\theta P_t)\,V_{t+1},
\qquad q_t=q_0P_0\cdots P_{t-1},\qquad V_{t+1}=P_{t+1}\cdots P_{K-1}r,
\]
with $V_K=r$; $q_t$ is the law of $X_t$, and $V_{t+1}(s)=\E[r(X_K)\mid X_{t+1}=s]$. The MH kernel is $P_t(x,\cdot)=\sum_yQ_t(y\mid x)\{\alpha_t(x,y)\delta_y+(1-\alpha_t(x,y))\delta_x\}$, and $Q_t$ does not depend on $\theta$, so
\[
\bigl[(\nabla_\theta P_t)V_{t+1}\bigr](x)=\sum_yQ_t(y\mid x)\,\nabla_\theta\alpha_t(x,y)\,\{V_{t+1}(y)-V_{t+1}(x)\}.
\]
Taking the expectation over $X_t\sim q_t$ and $Y_t\sim Q_t(\cdot\mid X_t)$ gives \eqref{eq:identity}. Self-proposals contribute nothing, because $V_{t+1}(y)-V_{t+1}(x)=0$ when $y=x$. Only the local acceptance probability is differentiated in each term; the effect of a decision on later steps enters through the value difference.

\paragraph{Paired continuations.}
Given $(X_t,Y_t)$, let $R^+$ and $R^-$ be the terminal rewards of continuations from $Y_t$ and $X_t$, each following the chain's transition law from step $t+1$ onward under the current energy and schedule. Then $\E[R^+-R^-\mid X_t,Y_t]=V_{t+1}(Y_t)-V_{t+1}(X_t)$ for any joint coupling of the two continuations; sharing random numbers changes only their joint law. By the Markov property, the base chain's own continuation from its realized $X_{t+1}$ has the correct conditional law, so it can serve as one branch, provided that whether step $t$ is selected does not depend on the base chain's future randomness. Shared uniforms pass through each branch's categorical distribution, and branches merge only when their states and all carried sampler quantities coincide, after which they agree. Continuations keep their original step indices, so they use the remaining part of the schedule.

\paragraph{Subsampling.}
Let $I_{it}$ indicate that decision $t$ of base chain $i$ is selected in the current update, with $\Pr(I_{it}=1\mid\mathcal F_{it})=p_{it}$, where $\mathcal F_{it}$ contains the prefix $X_{i,0:t}$, the proposal $Y_{it}$, and quantities fixed before the update, and $I_{it}$ is independent of the continuation randomness given $\mathcal F_{it}$. If $p_{it}>0$ whenever $\nabla_\theta\alpha_t(X_{it},Y_{it})\neq0$, then
\[
\E\Bigl[\frac{I_{it}}{p_{it}}(R^+_{it}-R^-_{it})\,\nabla_\theta\alpha_t(X_{it},Y_{it})\Bigr]
=\E\bigl[\{V_{t+1}(Y_t)-V_{t+1}(X_t)\}\,\nabla_\theta\alpha_t(X_t,Y_t)\bigr],
\]
with unselected terms set to zero. Summing over $t$ and averaging over the $N$ independent base chains shows that \eqref{eq:estimator} is unbiased for $\nabla_\theta J$. Dividing by the number of selected decisions instead of $N$ would introduce a bias. States, rewards, and inclusion probabilities are treated as constants during differentiation, so no gradient flows through sampled categorical states, inclusion weights, or the discriminator.

\paragraph{Regularity.}
Write $\Delta_t=\beta_t[E_\theta(Y_t)-E_\theta(X_t)]-\log\{Q_t(X_t\mid Y_t)/Q_t(Y_t\mid X_t)\}$, so that $\alpha_t=\min\{1,e^{-\Delta_t}\}$. If $\Delta_t>0$, then $\nabla_\theta\alpha_t=-\beta_t\alpha_t\nabla_\theta[E_\theta(Y_t)-E_\theta(X_t)]$, so increasing the proposed state's energy relative to the current one lowers acceptance. If $\Delta_t<0$, the proposal is strictly saturated and $\nabla_\theta\alpha_t=0$; if the reverse proposal probability is zero, $\alpha_t=0$ and its derivative vanishes. At a tie, $\Delta_t=0$ with $\nabla_\theta\Delta_t\neq0$, $\alpha_t$ is not differentiable; the implementation uses zero there, and no unbiasedness claim is made at ties. The statement assumes exact arithmetic and a fixed discriminator. It does not extend to floating-point effects, gradient clipping, coordinate constraints, adaptive optimization, discriminator refitting, or the alternating procedure as a whole, and it implies nothing about convergence or mixing.

Finite-state checks cover 96 configurations with asymmetric proposals, self moves, varying schedules, and the tie convention. Eleven deliberately injected implementation faults were detected where applicable. Additional inclusion-weight and coordinate checks agree with enumerated gradients to approximately $10^{-13}$ at regular points.

\subsection{Molecular Optimization}
The coefficients in \eqref{eq:energy} are written in energy coordinates $(u,w)$ as $c=a_0e^u$ and $\omega=we^{-u}$, so that $E_\theta=a_0e^u\VG+a_0w(\VG-\Vpre)$, starting at $(u,w)=(0,0)$, which is the calibrated baseline. The initial scales are $a_0=0.70,0.70,0.8649$ at $K=125,250,1000$ (rounded to 0.865 in the main text). The learned energy is kept unchanged when the acceptance rule changes (Table~\ref{tab:released}); coefficients are not rescaled using another setting's calibration.

AdamW operates on the reparameterized coordinates $(u',w)$, where $u'=\log(e^u+0.57w)$, so that $E_\theta=a_0[e^{u'}\VG+w(0.43\VG-\Vpre)]$. The feasible set is $u\in[-1.5,1.5]$, $w\in[-1,3]$, and $e^u+0.57w\ge e^{-1.5}$, so the argument of the logarithm defining $u'$ is at least $e^{-1.5}$. After each update, $u$ and $w$ are clipped to their intervals, and $w$ is then raised where necessary to satisfy the last constraint. The discriminator logit is $-\bref(V_\psi-\VG)$, where $V_\psi$ is the LoRA-adapted potential and $\bref=17.619552679597852$ is fixed to the final inverse temperature of GEM's released acceptance schedule; it is held constant across sampling budgets and discriminator updates.

\begin{table}[ht]
\centering
\caption{Molecular training configuration. Both pretrained potentials are frozen; only the two energy coordinates parameterize the acceptance energy. The discriminator is used only during training.}
\label{tab:training}
\small
\renewcommand{\arraystretch}{1.12}
\begin{tabular}{@{}>{\raggedright\arraybackslash}p{.25\textwidth}>{\raggedright\arraybackslash}p{.70\textwidth}@{}}
\toprule
Component & Setting \\
\midrule
Energy updates & 80 updates; 256 base chains per update; approximately four selected decisions per chain on average; final iterate.\\
Decision selection & Inclusion probability $\min(1,\lambda_j h_{it})$, with $h_{it}$ an acceptance-gradient norm in rescaled coordinates and guidance weight three. $\lambda_0=0.0501$; later scales $\lambda_j$ use the preceding update's sensitivity sum.\\
Optimizer & AdamW, learning rate 0.02, five-update warm-up then linear decay to zero, moments $(0.9,0.999)$, no weight decay.\\
Coordinates & AdamW on $(u',w)$ with $u'=\log(e^u+0.57w)$; gradient components divided by $(0.676,0.152)$ before clipping to norm one. After each update, clip $u\in[-1.5,1.5]$ and $w\in[-1,3]$, then raise $w$ where necessary so that $e^u+0.57w\ge e^{-1.5}$.\\
Discriminator & Rank-64 LoRA adaptation of $\VG$ on all 12 layers and the scalar head \citep{hu2022lora}; approximately 9.7 million trainable parameters; logit $-\bref(V_\psi-\VG)$ with fixed $\bref\approx17.62$.\\
Initial discriminator fit & 25,600 generated examples; 1,000 logistic updates with 128 generated and 128 node-count-matched data graphs per update. AdamW learning rate $2\times10^{-5}$, 200-step warm-up, no weight decay, clipping to one.\\
Refitting & One discriminator step after each energy update, reusing the first 128 of the 256 base-chain terminal graphs generated before that update, together with 128 node-count-matched training graphs.\\
\bottomrule
\end{tabular}
\end{table}

At $K=250$, discriminator negatives are calibrated GEM samples. Invalid negatives receive weight three in the logistic loss, selected from $\{1,3,9,27\}$ by held-out classifier loss. At $K=1000$, half the negatives are calibrated samples and half come from guidance settings $(c,\omega)=(0.5,1),(0.8649,1),(0.5,0.5)$; the invalid weight is fixed to three. The $K=125$ pool also mixes calibrated and guided samples. The three $K=1000$ discriminator seeds share their negative pool. Gradient clipping is active in about 85\% of the $K=250$ updates. The exact-state optimizer does not use these molecular clipping and coordinate settings.

\subsection{Alternative Estimators}
\label{app:controls}
A matched-transition experiment uses all six original exact targets, $K=4,8,16$, three fresh seeds, and the same 2,000-update schedule. Coupled learning uses 512 base chains; leave-one-out score-function learning uses 704, 736, and 752 chains, respectively, so that both simulate the same number of transitions. Both improve on the correct energy in all 18 settings. Coupled learning has lower median TV in 12 settings, with median difference $-0.0015$, but takes 4.66 times the CPU runtime. Learning rates were not separately tuned. A fixed-gradient diagnostic likewise establishes no general efficiency advantage over leave-one-out scores or paired finite differences. These findings support terminal learning, not superiority of coupling.

\subsection{Computational Resources and Continuation Work}
\label{app:resources}
Molecular energy training uses NVIDIA H100 GPUs: energy updates take approximately 4.86 GPU-hours per seed at $K=250$ and 16.3 at $K=1000$, excluding discriminator preparation. At $K=250$, generation takes 0.859 H100 GPU-seconds per requested graph for calibrated GEM and 0.932 for our two-potential energy, excluding packed batches. Main comparisons match sampling steps, not elapsed time or runtime at equal quality. For fixed graph size and bounded retries, both samplers require work proportional to $K$ beyond initialization.

Let $\mathcal S$ contain the selected decisions. Simulating the $N$ base chains and one full alternative continuation per selected decision uses at most
\[
NK+\sum_{(i,t)\in\mathcal S}(K-t-1)
\]
refinement transitions, excluding initialization, derivative evaluations, and discriminator work; earlier branch merging can lower this count. Branching at every step gives $NK+NK(K-1)/2$ transitions; selecting an expected $F$ decisions per chain gives at most $NK+NF(K-1)$ in expectation. With one uniformly selected step, as in the exact study, the expected count is $N(3K-1)/2$. These are counts of simulated transitions, not measured wall-clock ratios, and they do not bound mixing time. Molecular training targets about four decisions per chain. At deployment no continuations or discriminator evaluations are needed.

\section{Finite-State Experiments}
\label{app:exact}

\subsection{Targets and Optimization}
\label{app:exact-targets}
The finite-state study uses no external dataset and is separate from the molecular experiments: each target is a synthetic pairwise binary (Ising) model $p(s)\propto e^{-H(s)}$ over configurations $s=(s_1,\ldots,s_L)$ with $L=6$ or $L=8$ variables. Here $H(s)=-\sum_i h_i s_i-\sum_{i<j}J_{ij}s_i s_j$, with $s_i\in\{-1,1\}$, $h_i\sim\mathcal N(0,0.25^2)$, and $J_{ij}\sim\mathcal N(0,1.25^2/L)$; the fields $h_i$ bias individual variables, and the couplings $J_{ij}$ create dependencies between them. The reference and new-target sets each contain three targets of each size, drawn independently from this distribution, each with its own fields and couplings. Target IDs such as \texttt{Ref-6-1} identify these draws by set, number of spins, and instance index; they do not denote methods, difficulty levels, or training seeds. Enumeration gives target probabilities, initial marginals, and transition laws. The initial law is the product of target marginals; proposals flip one uniformly chosen spin, and rejection supplies self transitions. The inverse temperature rises from 0.2 to 1 over the first $\lceil K/2\rceil$ steps, including both endpoints, and is then fixed.

A restricted unary-and-ring energy is fitted by population maximum likelihood. Its scalar multiplier minimizes terminal TV over 81 logarithmically spaced values from 0.05 to 20. The learned family adds all unary and pairwise features, with 21 or 36 coefficients. Population pseudo-likelihood recovers the correct energy $H$ up to a constant \citep{besag1975pseudolikelihood}. The discriminator is linear in those features and an intercept, initialized by the exact population logistic fit and subsequently updated from generated and target samples.

The original learning grid uses $K\in\{4,8,16,32,64,128,256\}$, trajectory counts $N\in\{32,128,512\}$, oracle or fitted discriminators, coupled or independent continuations, and three seeds. All runs use 2,000 Adam updates, learning rate 0.02 decaying to zero, moments $(0.9,0.999)$, no clipping or weight decay, and final-iterate evaluation. Each chain selects one uniformly chosen decision. The discriminator receives at most 128 generated and 128 target samples per update.

The numerical reference minimizes $D(q\|p)=2\KL(\tfrac12(p+q)\,\|\,p)$ over the learned family, with $q=q_{\theta,K}$. Consider the balanced, unweighted population problem of classifying samples of $p$ against samples of $q$, and let $d^*$ be its optimal unrestricted discriminator, so that $\sigma(d^*)=p/(p+q)$. Hold $d^*$ fixed at the current generator, and assume $p>0$ wherever $q>0$, which holds here because every target $p\propto e^{-H}$ has full support. Using $\sum_x\nabla_\theta q(x)=0$,
\[
\nabla_\theta\E_q[\log\sigma(d^*)]=\sum_x\nabla_\theta q(x)\log\frac{p(x)}{p(x)+q(x)}=-\nabla_\theta D(q\|p),
\]
where $d^*$ is not differentiated. The reward gradient for this ideal, frozen discriminator is therefore the negative gradient of $D$. The relation describes the finite-state setting; it does not imply that the weighted, finitely trained molecular discriminator optimizes the same divergence. It is also separate from the estimator of Section~\ref{sec:estimator}, which is unbiased for $\nabla_\theta J$ with any fixed discriminator under the conditions stated there. The reference uses exact gradients, a coefficient box of $[-20,20]$, two starts, bounded L-BFGS-B, and projected-Adam refinement. It is a numerical reference rather than a certified global solution. Eight of 42 reported reference configurations fail the projected-gradient tolerance, mainly at acceptance kinks; additional minimizations and directional checks change the objective by at most about $7.3\times10^{-11}$. The TV recovery fraction divides the learned improvement over the calibrated restricted energy by the reference's improvement. Because the reference minimizes $D$, not TV, recovery does not measure a known maximum TV gain.

\subsection{Repeated Training and Scalar Controls}
Fresh-seed replication uses seeds 1001--1003 on the reference targets; new-target training uses seeds 3001--3003. Both follow the 512-chain, 2,000-update procedure fixed before these runs, without selecting final iterates by TV. Table~\ref{tab:exact-detail} lists each target by its ID and horizon $K$, and reports our method's median and sample SD across three training seeds. Its reference block reports the fresh-seed replication, whereas Table~\ref{tab:exact-main} and Figure~\ref{fig:exact-learning} report the original runs; the replication's cross-target medians therefore differ slightly from Table~\ref{tab:exact-main}. The medians determine win counts. Horizons within a target are dependent, so these counts are descriptive.

The scalar control minimizes exact terminal TV for $\beta H$ over nonnegative scales, including zero and the infinite-scale downhill-only limit. Logarithmic grids of 513 to 4,097 points are refined around detected minima until the best TV values change by less than $10^{-9}$. The omitted large-$\beta$ tail is controlled to $10^{-10}$ using the smallest positive single-flip energy increase. This controls the tail, not global optimality inside the search interval. All 18 reference and 18 new-target searches stabilize. Learned medians beat tuned $\beta H$ in 14 of 18 reference settings in both the original runs and the fresh-seed replication, and in 15 of 18 new-target settings; the three new-target exceptions are all on \texttt{New-8-2}. Independent probability propagation verifies the terminal laws of these follow-up runs.

\begin{table}[p]
\centering
\caption{Exact terminal total variation by target distribution and sampling horizon; lower is better. Each target is a synthetic pairwise binary model with random fields and couplings (Appendix~\ref{app:exact-targets}); no external dataset is used. Target IDs specify the set (\texttt{Ref} or \texttt{New}), number of spins (6 or 8), and instance index (1--3). Each instance has fixed fields and couplings; its three rows evaluate the same target at horizons $K=4,8,16$, the number of MH transitions. The controls are the equilibrium-correct energy $H$ and its numerically tuned scalar multiple $\beta H$; because $H$ is exact at equilibrium, lower TV than $H$ reflects a finite-horizon improvement. Learned columns report the median and sample standard deviation of terminal TV across three training seeds; TV itself is evaluated exactly, so the SD reflects only variation across training seeds. The reference block repeats training with fresh seeds on the original targets, whereas the new-target block trains from scratch on independently drawn target distributions to test whether the gains recur; no trained energy is transferred. Table~\ref{tab:exact-main} and Figure~\ref{fig:exact-learning} use the original reference-target runs. All learned runs use $N=512$ and 2{,}000 updates. Bold identifies the lowest of the three TV values using unrounded results.}
\label{tab:exact-detail}
\small
\renewcommand{\arraystretch}{1.08}
\begin{tabular*}{.96\textwidth}{@{\extracolsep{\fill}}lrrrrr@{}}
\toprule
Target ID & $K$ & Correct $H$ & Tuned $\beta H$ & Learned median & Seed SD \\
\midrule
\multicolumn{6}{@{}l}{\textit{Reference targets: fresh-seed replication}} \\
\texttt{Ref-6-1} & 4 & $0.1266$ & $0.0613$ & $\mathbf{0.0529}$ & $0.005$ \\
\texttt{Ref-6-1} & 8 & $0.0845$ & $0.0284$ & $\mathbf{0.0275}$ & $0.004$ \\
\texttt{Ref-6-1} & 16 & $0.0593$ & $0.0279$ & $\mathbf{0.0208}$ & $0.004$ \\
\addlinespace[2pt]
\texttt{Ref-6-2} & 4 & $0.3435$ & $0.2319$ & $\mathbf{0.2284}$ & $0.001$ \\
\texttt{Ref-6-2} & 8 & $0.2389$ & $0.1247$ & $\mathbf{0.0805}$ & $0.009$ \\
\texttt{Ref-6-2} & 16 & $0.1234$ & $0.0524$ & $\mathbf{0.0359}$ & $0.005$ \\
\addlinespace[2pt]
\texttt{Ref-6-3} & 4 & $0.3999$ & $0.3126$ & $\mathbf{0.3045}$ & $<0.001$ \\
\texttt{Ref-6-3} & 8 & $0.2326$ & $0.1041$ & $\mathbf{0.0810}$ & $0.024$ \\
\texttt{Ref-6-3} & 16 & $0.0838$ & $\mathbf{0.0242}$ & $0.0759$ & $0.007$ \\
\addlinespace[2pt]
\texttt{Ref-8-1} & 4 & $0.2672$ & $0.1803$ & $\mathbf{0.1534}$ & $0.002$ \\
\texttt{Ref-8-1} & 8 & $0.1549$ & $0.0917$ & $\mathbf{0.0435}$ & $0.003$ \\
\texttt{Ref-8-1} & 16 & $0.0935$ & $0.0633$ & $\mathbf{0.0247}$ & $0.004$ \\
\addlinespace[2pt]
\texttt{Ref-8-2} & 4 & $0.3142$ & $0.2235$ & $\mathbf{0.2095}$ & $<0.001$ \\
\texttt{Ref-8-2} & 8 & $0.1686$ & $0.0914$ & $\mathbf{0.0496}$ & $0.003$ \\
\texttt{Ref-8-2} & 16 & $0.0612$ & $0.0369$ & $\mathbf{0.0318}$ & $0.011$ \\
\addlinespace[2pt]
\texttt{Ref-8-3} & 4 & $0.3701$ & $\mathbf{0.1807}$ & $0.1844$ & $<0.001$ \\
\texttt{Ref-8-3} & 8 & $0.3012$ & $\mathbf{0.0939}$ & $0.1174$ & $<0.001$ \\
\texttt{Ref-8-3} & 16 & $0.2198$ & $\mathbf{0.0636}$ & $0.0923$ & $0.007$ \\
\addlinespace[2pt]
\midrule
\multicolumn{6}{@{}l}{\textit{New targets: separately trained}} \\
\texttt{New-6-1} & 4 & $0.6815$ & $0.6149$ & $\mathbf{0.6106}$ & $<0.001$ \\
\texttt{New-6-1} & 8 & $0.4960$ & $0.3693$ & $\mathbf{0.3459}$ & $0.002$ \\
\texttt{New-6-1} & 16 & $0.2572$ & $0.1132$ & $\mathbf{0.0746}$ & $0.008$ \\
\addlinespace[2pt]
\texttt{New-6-2} & 4 & $0.1657$ & $0.0762$ & $\mathbf{0.0434}$ & $0.002$ \\
\texttt{New-6-2} & 8 & $0.0977$ & $0.0469$ & $\mathbf{0.0338}$ & $0.002$ \\
\texttt{New-6-2} & 16 & $0.0658$ & $0.0448$ & $\mathbf{0.0160}$ & $0.005$ \\
\addlinespace[2pt]
\texttt{New-6-3} & 4 & $0.4313$ & $0.3193$ & $\mathbf{0.2996}$ & $0.002$ \\
\texttt{New-6-3} & 8 & $0.2894$ & $0.1612$ & $\mathbf{0.1241}$ & $0.005$ \\
\texttt{New-6-3} & 16 & $0.1416$ & $0.0432$ & $\mathbf{0.0319}$ & $0.002$ \\
\addlinespace[2pt]
\texttt{New-8-1} & 4 & $0.5173$ & $0.4715$ & $\mathbf{0.4711}$ & $<0.001$ \\
\texttt{New-8-1} & 8 & $0.3651$ & $0.3172$ & $\mathbf{0.2677}$ & $<0.001$ \\
\texttt{New-8-1} & 16 & $0.2224$ & $0.1898$ & $\mathbf{0.0643}$ & $0.002$ \\
\addlinespace[2pt]
\texttt{New-8-2} & 4 & $0.5886$ & $\mathbf{0.5116}$ & $0.5655$ & $0.002$ \\
\texttt{New-8-2} & 8 & $0.4601$ & $\mathbf{0.3307}$ & $0.4097$ & $0.002$ \\
\texttt{New-8-2} & 16 & $0.2856$ & $\mathbf{0.1180}$ & $0.2553$ & $0.002$ \\
\addlinespace[2pt]
\texttt{New-8-3} & 4 & $0.6175$ & $0.5830$ & $\mathbf{0.5716}$ & $<0.001$ \\
\texttt{New-8-3} & 8 & $0.4878$ & $0.4123$ & $\mathbf{0.3664}$ & $0.003$ \\
\texttt{New-8-3} & 16 & $0.3019$ & $0.1852$ & $\mathbf{0.0968}$ & $0.002$ \\
\addlinespace[2pt]
\bottomrule
\end{tabular*}

\end{table}

\subsection{Finite-Time and Equilibrium Distributions}
\label{app:example}
Let the target be uniform on $\{a,b\}$, start at $a$, and propose a flip with probability $3/4$. The correct energy accepts every flip, giving probability $3/4$ to $b$ after one step. At inverse temperature one, setting $E(b)-E(a)=\log(3/2)$ reduces acceptance to $2/3$: the one-step output is uniform, although the equilibrium probability of $b$ is $2/5$.

\begin{figure}[ht]
\centering
\resizebox{.60\textwidth}{!}{\definecolor{ovBlue}{HTML}{0072B2}
\begin{tikzpicture}[x=1cm,y=1cm,font=\scriptsize,
 every node/.style={inner sep=1.5pt},>=Stealth]
\node[anchor=south] at (4.05,2.58) {One-step output};
\node[align=center,anchor=south] at (7.05,2.58) {Equilibrium\\$\Pr(b)$};
\node[anchor=east,align=right] at (2.30,2.13) {Target};
\node[anchor=east,align=right] at (2.30,1.29) {Equilibrium-\\correct energy};
\node[anchor=east,align=right] at (2.30,.45) {Finite-time\\correction};
\filldraw[fill=black!10,draw=black!60,line width=.45pt] (2.55,1.89) rectangle (4.05,2.37);
\filldraw[fill=ovBlue!20,draw=black!60,line width=.45pt] (4.05,1.89) rectangle (5.55,2.37);
\node at (3.30,2.13) {$a:\,\frac12$};\node at (4.80,2.13) {$b:\,\frac12$};
\node at (7.05,2.13) {$\frac12$};
\filldraw[fill=black!10,draw=black!60,line width=.45pt] (2.55,1.05) rectangle (3.30,1.53);
\filldraw[fill=ovBlue!20,draw=black!60,line width=.45pt] (3.30,1.05) rectangle (5.55,1.53);
\node at (2.925,1.29) {$a:\,\frac14$};\node at (4.425,1.29) {$b:\,\frac34$};
\node at (7.05,1.29) {$\frac12$};
\filldraw[fill=black!10,draw=black!60,line width=.45pt] (2.55,.21) rectangle (4.05,.69);
\filldraw[fill=ovBlue!20,draw=black!60,line width=.45pt] (4.05,.21) rectangle (5.55,.69);
\node at (3.30,.45) {$a:\,\frac12$};\node at (4.80,.45) {$b:\,\frac12$};
\node at (7.05,.45) {$\frac25$};
\draw[black!25,line width=.5pt] (6.0,.07)--(6.0,2.85);
\end{tikzpicture}}
\caption{\textbf{Finite-time accuracy need not imply equilibrium accuracy.} One-step output and equilibrium probability for the two-state example, computed exactly.}
\label{fig:example}
\end{figure}

\FloatBarrier
\section{Molecular Evaluation}
\label{app:molecular}

\subsection{Sampler, Calibration, and Metrics}
Each chain draws 8--27 nodes according to the MOSES training histogram and keeps this count fixed. Initialization performs at most 225 greedy edits under $\VG$. We retain the released proposal settings, approximately $\beta_{\mathrm{prop}}=2.36$, $\lambda_X=4.00$, and $\lambda_E=4.83$. If no label changes, the sampler retries after multiplying these settings by 0.9, for at most 20 stages. Acceptance uses an inverse-temperature ramp from approximately 0.78 to 17.62 over 500 steps, then holds the final value. Continuations retain their original step index. Values here are rounded for readability; full-precision proposal and acceptance-temperature constants are provided in \texttt{molecular.py} in the supplementary code, and its YAML file specifies experiment settings. The molecular example in the supplement implements the training method but does not reconstruct the original development, selection, and final evaluation pipeline.

Exact MH marginalizes both forward and reverse proposal probabilities over the same 20 stages. GEM's released rule instead uses the realized forward retry path and a reverse marginal with up to 256 stages and log-mass floor $-50$; it is not exact MH for the same target. Both rules have parameter-independent proposal terms. Network calls are padded to 256 graphs; sampler precision follows the release, while discriminator scoring and refitting use the highest float32 precision.

Development FCD, used for baseline calibration, discriminator screening, and static-coefficient selection, compares 2,048 generated graphs per candidate setting, from a generation stream separate from the final comparisons, with a fixed reference of 16,010 molecules: the MOSES training molecules whose SHA-256 hash of the canonical SMILES is divisible by 100 (about 1\% of the training set). Final FCD uses the separate MOSES scaffold-test reference of 176,225 molecules. The development molecules were not held out from the pretraining of the released GEM checkpoints. The following calibration scales are rounded to three decimal places. Exact GEM selects 0.7 at $K=250$ from $\{0.25,0.35,0.5,0.7,0.865,1,1.5,2,3\}$, a grid extended after an initial boundary solution, and 0.865 at $K=1000$ from $\{0.5,0.865,1\}$. At $K=125$, the selected 0.7 nearly ties 0.865 in development FCD. Released-rule GEM at $K=250$ selects 0.5 over 0.7 and 1; the first two differ by only 0.0013 in FCD but substantially in validity. The released baselines at $K=125,1000$ use scale one, not a budget-calibrated scale.

Every requested graph counts in the denominators of validity and VUN, and failed outputs are not replaced. Native validity permits disconnected sanitizable molecules; native FCD uses their canonical largest fragments, retaining duplicates. Strict validity requires connectedness and the stated canonical representation checks. Novelty uses the unfiltered training set of 1,584,663 canonical SMILES. Software is Python 3.10.21, PyTorch 2.5.1 with CUDA 12.1, RDKit 2026.03.6, \texttt{fcd\_torch} 1.0.7, NumPy 2.2.6, and SciPy 1.15.2. Scoring runs on CPUs. The graph encoding cannot represent every aromatic-hydrogen configuration, so these scores are interpreted within the same encoding and evaluator rather than against published leaderboard values.

\subsection{Released-Rule Deployment and Uncertainty}
\begin{table}[ht]
\centering
\caption{\textbf{Deployment with GEM's released acceptance rule.} ``GEM + learned energy (Ours)'' reuses the energies from Table~\ref{tab:molecular} without retraining, with released-rule acceptance replacing exact MH. Metrics and denominators match Table~\ref{tab:molecular}: FCD, validity (\%), and VUN (\%), with 25,000 requested graphs per seed. Entries are means $\pm$ sample SD over two prespecified seeds at $K=125$ and three elsewhere. Baselines are uncalibrated released GEM at $K=125,1000$ and calibrated GEM at $K=250$. Bold marks the better mean within each pair, not statistical significance.}
\label{tab:released}
\renewcommand{\arraystretch}{1.12}
\begin{tabular*}{.98\textwidth}{@{\extracolsep{\fill}}rlrrr@{}}
\toprule
$K$ & Method & FCD $\downarrow$ & Validity $\uparrow$ & VUN $\uparrow$ \\
\midrule
125 & GEM (released) & $2.406\pm0.0003$ & $\mathbf{93.25\pm0.05}$ & $\mathbf{90.99\pm0.06}$ \\
 & GEM + learned energy (Ours) & $\mathbf{2.295\pm0.011}$ & $92.45\pm0.44$ & $90.43\pm0.44$ \\
\midrule
250 & GEM (calibrated) & $1.920\pm0.012$ & $93.48\pm0.09$ & $91.20\pm0.10$ \\
 & GEM + learned energy (Ours) & $\mathbf{1.784\pm0.033}$ & $\mathbf{95.02\pm0.42}$ & $\mathbf{92.97\pm0.41}$ \\
\midrule
1000 & GEM (released) & $1.529\pm0.013$ & $\mathbf{98.02\pm0.03}$ & $94.02\pm0.22$ \\
 & GEM + learned energy (Ours) & $\mathbf{1.487\pm0.022}$ & $97.95\pm0.08$ & $\mathbf{94.21\pm0.11}$ \\
\bottomrule
\end{tabular*}

\end{table}

The internally prespecified decision rule combines a one-sided paired FCD test at level 0.05 with one-sided 95\% lower confidence limits above $-1$ percentage point for the validity and VUN differences, each computed as learned minus baseline. The exact-rule comparison meets it at $K=125,1000$; at $K=250$, the validity lower limit of $-1.25$ points does not establish non-inferiority. Released-rule deployment meets it at $K=250,1000$, not at $K=125$, where the validity and VUN lower limits are $-2.52$ and $-2.80$ points. A later three-seed analysis also fails those conditions and does not replace the prespecified two-seed result. No multiplicity adjustment is applied across the six comparisons.

Training seed $k$ is paired with generation seed $1911+k$. The two or three seed-level differences, not individual graphs, are the observations for the small-sample $t$ tests. Energies are reused across acceptance rules, and some discriminator data are shared.

Learned rows use H100 GPUs. One released-GEM $K=1000$ comparator seed mixes 63 H100 and 35 L40-class batches; the two H100-only seeds give an unresolved FCD comparison ($p=0.11$). All $K=125$ rows use H100s. Runtime estimates exclude packed batches.

\subsection{Protocol Changes and Sensitivity}
\label{app:molecular-protocol}
Protocols were recorded internally, not in an external registry. The original $K=250$ comparison followed static development results and exploratory pilots but preceded final learned outcomes. Its released-rule hypothesis followed development results, and later budgets followed earlier tests. The six comparisons are therefore not independent prospective replications. No final test score selected the deployed iterate.

At $K=125$, a baseline-only encoding check found 22 mismatches among 1,833 native-valid graphs (1.20\%). The discriminator-pool eligibility threshold was raised from 1\% to 2\% before any pool output or discriminator training. The completed pool contained 368 mismatches among 22,270 valid graphs (1.65\%): 322 non-isomorphic re-encodings and 46 explicit-hydrogen cases. The change affected eligibility, not evaluation or sample handling. The original rule would have stopped the experiment.

Strict scoring preserves the exact-rule $K=125$ improvement: validity and VUN gains are 4.77 and 4.65 points. About 0.64 points of the native validity gain arise from disconnected output. Matching the valid-input counts changes the FCD difference from $-0.1419$ to $-0.1379$; all fixed subsample comparisons remain negative. These checks reuse generated samples and do not establish that encoding mismatches in the discriminator's training pool are harmless. The released-rule $K=125$ joint failure also remains under strict scoring.

At $K=1000$, the initial discriminator failed a screen relating reward to development FCD. A revised negative pool containing guided samples passed at the correlation boundary of $-0.50$. Several screened settings also supplied negatives, and all three discriminator seeds shared the pool. At $K=500$, the screen failed with correlation approximately $-0.071$; no energy was trained and no second attempt was made. The absence of that budget is not an evaluated negative generator result.

\subsection{Scope of the GEM Comparison}
\label{app:molecular-controls}
Coefficient-selection controls give FCD 2.280 at $K=125$, versus 2.366 for learning, but lower validity (91.10\% versus 92.55\%); the FCD difference is unresolved with three seeds. At $K=250$, their FCD is 1.811 versus 1.829 with exact acceptance, and 1.769 versus 1.784 with released acceptance; rate differences are small. Finite-difference training gives exact-rule FCD 1.821 under a different optimization schedule. These comparisons do not establish an FCD advantage for learning over the alternatives; we report only these summaries.

The $K=125$ static setting was selected among four development candidates. The $K=250$ setting has the lowest development FCD among tested settings but is validity-ineligible: 94.97\% against a 95.44\% threshold. These descriptive controls use three seeds with shared generation streams; there is no corresponding three-seed static evaluation at $K=1000$.

The six-setting coefficient grid costs 3.2 H100-hours at $K=250$, versus 4.86 per seed for energy updates before discriminator preparation. Schedule compression is inconclusive, and cross-budget transfer is confounded by the change of scale. A larger residual energy did not improve development FCD at matched scales; its planned tests and a random-sign placebo test were not completed. Molecular-weight distance worsens at $K=1000$ despite improved FCD, precluding a claim of uniformly improved molecular properties.

\end{document}